\documentclass[journal]{IEEEtran}

\usepackage{url}
\usepackage{scicite} 
\usepackage{amsmath} 
\usepackage{amssymb}
\usepackage{bm}
\usepackage{acro}
\usepackage{balance}
\usepackage{booktabs}

\usepackage{pifont}
\usepackage{color}
\usepackage{placeins}
\usepackage[]{graphicx} 
\usepackage{hyperref}
\usepackage{times}
\usepackage{tabularx} 
\usepackage{subcaption}
\usepackage{xcolor}
\definecolor{color2}{rgb}{0,0,0}
\usepackage{enumitem}
\usepackage[TS1,T1]{fontenc}

\begin{document}

\title{Robot builds a robot's brain: AI generated drone command and control station hosted in the sky}

\author{Peter J. Burke%
\thanks{Peter J. Burke is with the Department of Electrical Engineering and Computer Science, University of California, Irvine, CA 92697 USA (e-mail: pburke@uci.edu).}
}

\maketitle

\begin{abstract}

Advances in artificial intelligence (AI) including large language models (LLMs) and hybrid reasoning models present an opportunity to reimagine how autonomous robots such as drones are designed, developed, and validated. 
Here, we demonstrate a fully AI-generated drone control system: with minimal human input, an artificial intelligence (AI) model authored all the code for a real-time, self-hosted drone command and control platform, which was deployed and demonstrated on a real drone in flight as well as a simulated virtual drone in the cloud. The system enables real-time mapping, flight telemetry, autonomous mission planning and execution, and safety protocols—all orchestrated through a web interface hosted directly on the drone itself. Not a single line of code was written by a human.
We quantitatively benchmark system performance, code complexity, and development speed against prior, human-coded architectures, finding that AI-generated code can deliver functionally complete command-and-control stacks at orders-of-magnitude faster development cycles, though with identifiable current limitations related to specific model context window and reasoning depth. Our analysis uncovers the practical boundaries of AI-driven robot control code generation at current model scales, as well as emergent strengths and failure modes in AI-generated robotics code. This work sets a precedent for the autonomous creation of robot control systems and, more broadly, suggests a new paradigm for robotics engineering—one in which future robots may be largely co-designed, developed, and verified by artificial intelligence. In this initial work, a robot built a robot's brain.

\end{abstract}

%

\section*{INTRODUCTION}

In Arnold Schwarzenegger's Terminator, the robots become self-aware and take over the world. In this paper, we take a first step in that direction: A robot (AI code writing machine) creates, from scratch, with minimal human input, the brain of another robot, a drone.  

\subsection*{Man vs. machine}
\begin{figure*}
     \centering
     \includegraphics[width=6in]{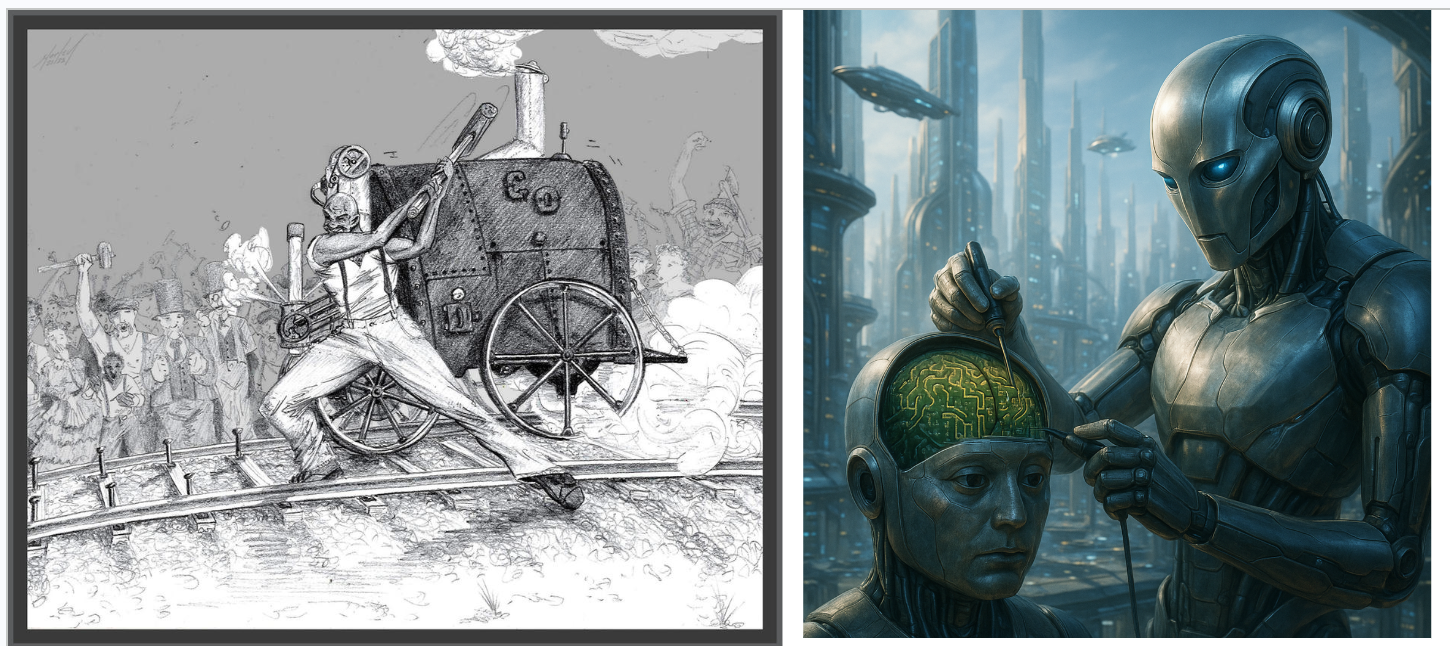}
     \caption{\textbf{Man vs. machine.} A) Legend has it John Henry tried to lay rail faster than a steam engine in the 1870s. He won, but died of exhaustion. (Illustration \textcopyright
harley‑1979 / deviantart.com/harley‑1979) B) A robot can program a robot's brain, demonstrated in this paper. 
     }
     \label{fig: manvsmachine}
\end{figure*}

 Legend has it that, in the 1870s, a human rail layer (John Henry) tried to beat a steam engine rail laying machine (robot) (Figure~\ref{fig: manvsmachine}A). He was extremely strong. He died trying to beat the machine (robot). John Henry is an American legend and icon, similar to Johny Appleseed, Paul Bunyan, and George Washington. The United States Postal Service issued a postage stamp of him in 1996. 
According to a folk song from 1918, later popularized by Disney, and still sung by American schoolchildren to this day, the American labor legend 'John Henry was a mighty man, born with a hammer right in his hand'\cite{kennedy1939johnhenry}. Although the machine lost to John Henry (just barely), for better or worse, the days of John Henry appear to be over. In this work, we demonstrate a similar result in robot control software.

 Figure~\ref{fig: manvsmachine}B summarizes the symbolic significance of this work. Although we will present some details of the particular "brain" created herein, the significance of the work is more to demonstrate the \emph{approach} the use of agentic AI to generate robot-controlling code \emph{from scratch}. This is distinct from, say, training a large language model (LLM) to provide optimum flight path for endurance or speed, approaches which use a drone brain already in existence and just tune the weighting factors. As such, this is an example of what could be accomplished in the future for a large variety of different kinds of "brains" for various tasks, using little to no human labor. The space of available robot applications is thus multiplied from scaling based on hours of human labor to scaling based on AI capacity, a quantum leap in the ability to create robots for new applications at the virtual snap of a finger.

\subsection*{A high level drone brain}
The human brain is divided into sections (both logical and anatomical/physical), some containing basic functions such as breathing and reflex, some containing higher level thought such as reasoning, and some intermediate level such as gait, and motor control. The analogy for a drone is as follows: In a drone, there is low-level firmware ("flight controller") such as Ardupilot\cite{ardupilot}, Pixhawk PX4\cite{px4} Betaflight\cite{betaflight}, etc. which runs at the base layer and sends commands directly to the motors, in order to keep the drone level without pilot input, for example. (Just like the human brain keeps the lungs breathing at night without input from the conscious mind). They interpret pilot commands from the position of fingers and joysticks on a remote control. This is like breathing and reflex in a human brain. It would be a larger project that AI may not yet be ready for to demonstrate a low-level flight controller firmware from AI \emph{from scratch}. Ardupilot has a million lines of code which is beyond the scope of AI to create from scratch as of this writing (summer 2025). In any case, such a low-level code requires extensive development and is not the scope of this work. 

A higher level computer, with on-board image processing and higher-level though process, is also not the topic of this paper. That "brain" is typically Robot Operating System (ROS\cite{ROS2}) or some other autonomous code with general goals such as collision avoidance, but not pilot controlled. ROS also has around a million lines of code. Creating an AI written version of ROS or equivalent is also probably not possible with current AI models, just like writing Linux from scratch is also not possible with current AI models. 

Rather, the scope of this work is the intermediate brain, which presents data in a uniform manner such as the drone position on a map, and allows higher-level commands to be sent to the drone, such as waypoint missions, auto- land, auto-take off, which is one level higher than the base firmware. Usually these functions are handled by a custom program off the drone called a ground control station (GCS). The "ground" in GCS means that it is usually run on a computer on the ground. Mission Planner\cite{missionplanner} and QGroundControl\cite{qgroundcontrol} are examples. They require a dedicated Linux, Mac, or Windows machine and use the windowing system of the operating system (OS). In this work, we have used AI to create a similar control software (from scratch), but with two key differences: 1) The control station is also a website host, accessible from any browser, meaning it can be run remotely from anywhere in the world, on any operating system, including tablets and smartphones, and 2) it can actually run on a computer on the drone (called a companion computer), making the drone a flying website. We and others have already demonstrated \#1 with extensive, manual coding (years of effort)\cite{cloudstation,cloudstation2024} and references therein. In fact, some entire companies (reviewed by us in~\cite{cloudstation,cloudstation2024}) are based on selling the code in 1), so it is not unreasonable to estimate that millions of dollars in human time (developers) have gone into this. In our experience, we have had three generations of undergraduate and graduate students develop increasingly sophisticated on a web-based ground control stations. One version of these was used to set a Guinness aviation record for the longest distance to pilot a drone (around the world)\cite{guinness}. To our knowledge, creating a website in they sky (\# 2) has not yet been demonstrated. The entire intermediate "brain" was coded in about 2 weeks of human time. The development process, function, testing, simulation, and flight demonstration is presented in this paper. We call it "WebGCS" (web ground control station) because it is web-based, although the website is hosted on the drone (in the air and not ground), so it is actually a website in the air.

\section*{RESULTS}
\subsection*{Ground based brain: Web Ground Control Station}

\begin{figure*}
    \centering
    \includegraphics[width=127mm]{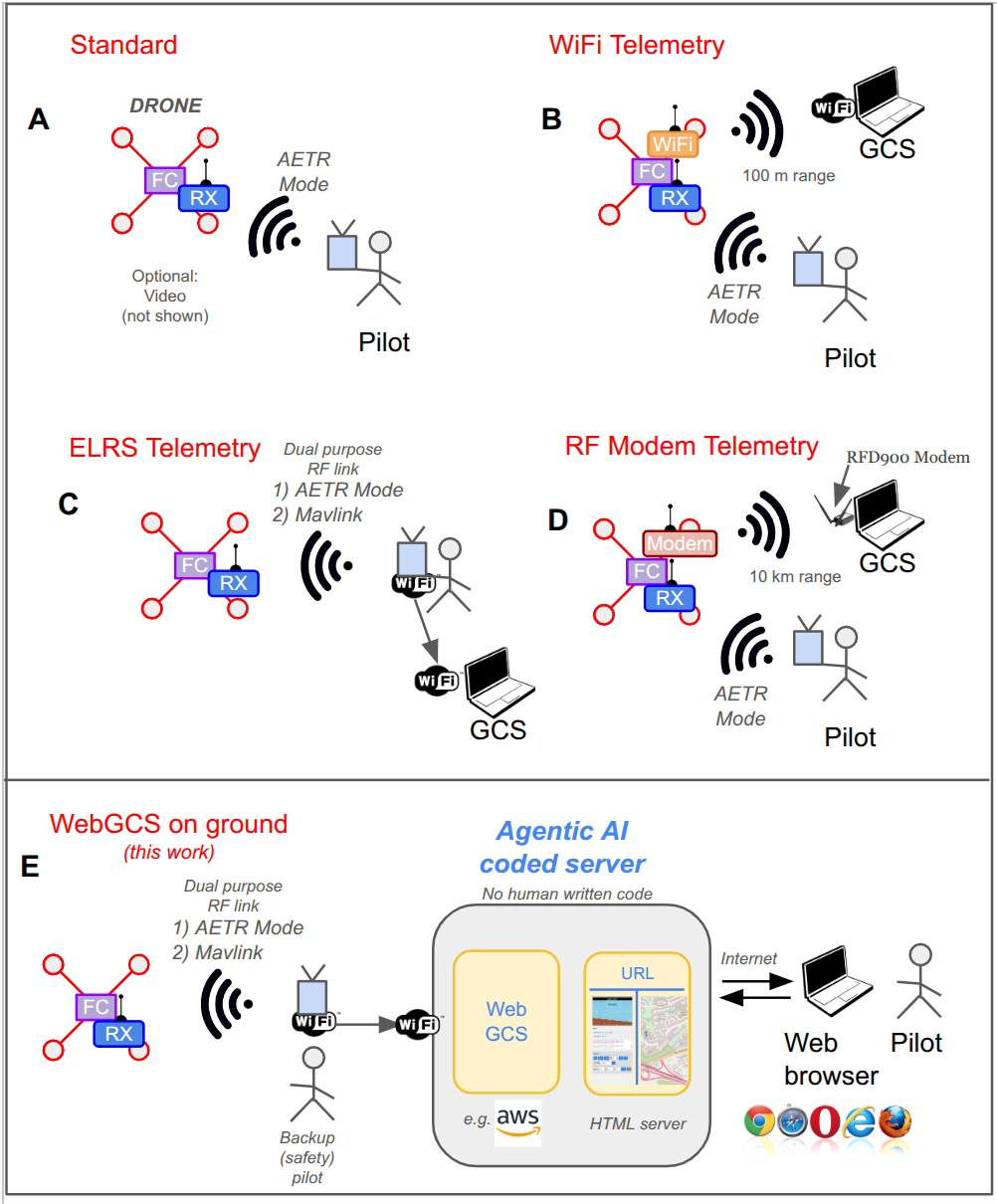}
    \caption{\textbf{Ground control stations.}
    (\textbf{A}) Standard setup: no GCS, only manual pilot control of aileron, elevator, rudder, throttle (AETR): "Stick and rudder".
    (\textbf{B}) WiFi telemetry 
    (\textbf{C}) ELRS telemetry
    (\textbf{D}) RF Modem Telemetry
    (\textbf{E}) Web ground control station coded entirely by AI (this work.) FC=flight controller. RX=receiver. See text for detailed explanation.
    }
    \label{fig: WebGCS}
\end{figure*}

The first major result of this work, the AI coded drone "brain", is on the ground. This is usually referred to as a ground control station (GCS), since it is a computer program on the ground that controls the drone in the air through a wireless telemetry link. Examples of this are Mission Planner and QGroundControl.
In Figure~\ref{fig: WebGCS}, we show the variety of drone wireless links in use today. In the standard "stick and rudder" method, a pilot uses a 4 channel radio (aileron elevator throttle rudder AETR) to manually pilot the drone, with an on-board radio receiver (RX) on the drone. In order to provide some automated computer control of the drone, a digital wireless link sending commands to the drone is used from a computer on the ground running GCS software, directly to the drone flight controller (FC) in the air, through some sort of modem. This can be a WiFi transceiver (Figure~\ref{fig: WebGCS}b, a second bit stream together with manual control (ExpressLRS Mavlink, Figure~\ref{fig: WebGCS}c, or a dedicated RF modem such as the RFD900\cite{rfd900-ardupilot}, Figure~\ref{fig: WebGCS}d. The common command set used by drones using the Ardupilot firmware is Mavlink\cite{mavlink}, a set of low- and high-level commands for drone command, control, communication, and telemetry.

In this work, AI is used to code a GCS on the ground shown in Figure~\ref{fig: WebGCS}e. This has the basic functionality of a GCS such as Mission Planner. However, the graphical user interface (GUI) is a website, and the website is also coded by the AI. To be clear, this is \emph{not} just a website, it includes a full stack ground control station. The website is just the interface to the backend. This means any pilot that can access the website hosted on the ground computer can remotely pilot the drone, a significant extension over existing GCS programs such as Mission Planner and QGroundControl, which require access to the terminal of the computer running the  custom GCS code. This adds additional flexibility of different operation systems for the pilot, and different locations. Finally, this adds the possibility to host the GCS in the cloud. (For example, with ELRS Mavlink the radio on the ground provides a WiFi connection, it can be connected to the internet.) For this reason, we name the AI coded ground control station "WebGCS". This is a robot brain on the ground, coded by a robot.

 \subsection*{Air based brain: Website in the sky}

The second major result of this work is a drone brain in the air. On a drone, higher level code is normally run on a so-called "companion computer", since it is the higher level "brain" function of the drone (analogous to the cerebral cortex in the human brain), leaving the lower level functionality (such as self-leveling and stability, analogous to the brain stem in the human brain controlling breathing) to the "low level" on board computer, the flight controller. 

\begin{figure*}
    \centering
    \includegraphics[width=185mm]{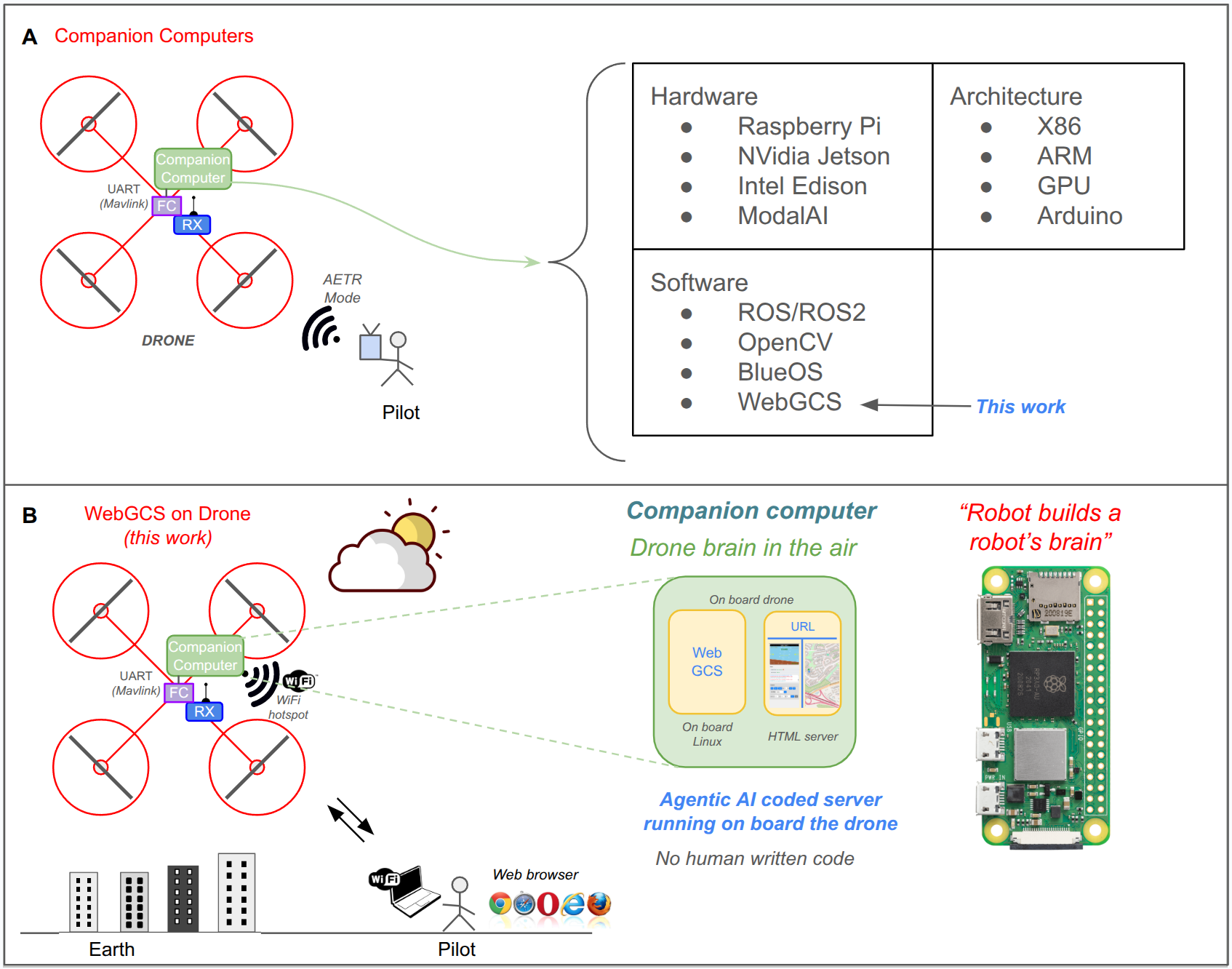}
    \caption{\textbf{On board computers.}
    (\textbf{A}) State of the art of companion computers (hardware/software) before this work. All software was written line by line by \emph{humans}.
    (\textbf{B}) A flying web based drone control system (this work), written entirely by \emph{machines}. 
    }
    \label{fig: onboardcomputers}
\end{figure*}

In Figure~\ref{fig: onboardcomputers}A, we show the existing state of the art of companion computers, both hardware, operating system, and software. The software (until this work) is essentially self contained on the drone, with communications to the cloud for additional functionality rare. Importantly, all of the existing software was "hand written", line by line, by \emph{humans}. For example, the most common companion computer codebase with higher level functionality is the so-called "Robot Operating System 2" \cite{ROS2}, which has over 200k lines of \emph{human} written code.

In this work, agentic AI large language models (LLMs) create an on board computer drone control station, which runs on a Linux machine on board the drone (in our case, a Raspberry Pi 2 W). The drone control station in the air has the same functionality of the ground control station in the first result, but it is in the air. In addition, in contrast to the existing software, it hosts it's own website \emph{in the air}, with no need for any special software on the ground other than a web browser (Figure~\ref{fig: onboardcomputers}B). All of the code is created by AI. Thus, in contrast to ROS2, and other drone control code in the air written by \emph{humans}, this drone control code was entirely created by a \emph{machine}: A robot brain (in the air) was created by a robot.

\subsection*{Pipeline: How a robot creates a robot brain}

\begin{figure*}
    \centering
    \includegraphics[width=185mm]{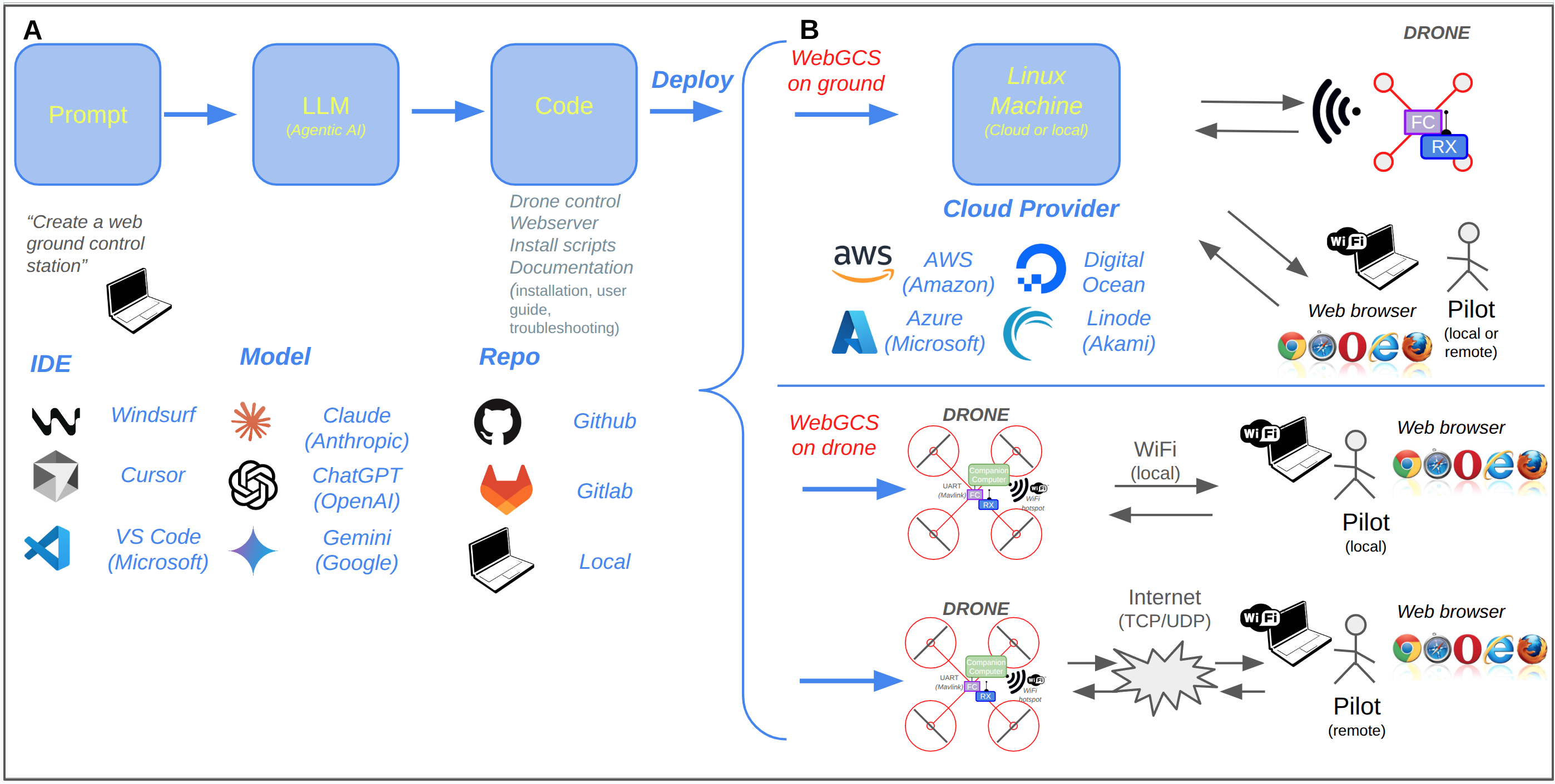}
    \caption{\textbf{Pipeline.}
    (\textbf{A}) Development pipeline.
    (\textbf{B}) Deployment pipeline.
    }
    \label{fig: pipeline}
\end{figure*}

We relied on several technologies to create a pipeline of innovation that ultimately resulted in an AI generated on board drone control station accessible as a website in the air (Figure~\ref{fig: pipeline}). The initial step is to set up an integrated development environment (IDE) which is compatible with AI coding, and also reasoning and planning, beyond just "hello world". The IDE is given the broad and specific goals, and executes them in generating code, with both local processing power in the AI LLM model contained with the IDE, as well as cloud based models available through a network connection. The codebase is synced with a cloud or local version control system (VCS) such as GitHub, GitLab, or git locally hosted. Some of the IDEs offer testing of various functionalities during the development process.

Once the code is sufficiently developed by the AI, it is then deployed to either a ground based control station (GCS), or a control station on the drone itself. When deployed on the ground, the code can run on a local machine or in the cloud, as long as an internet connection from the drone to the GCS is maintained. The pilot then uses a web browser to log into the WebGCS to control the drone. When deployed on the drone in the air, the drone hosts the website directly from the air. The pilot can connect (log into) the drone from a local WiFi connected PC as the drone actually creates a WiFi hotspot in the air. In addition, if desired, the TCP traffic can be routed to any computer in the world, where the pilot can connect and control the drone from.

\subsection*{Comparing man to machine: Vibe coding a drone brain}

Coming back to John Henry, in this section we describe in detail how we vibe-coded the drone brain, and then later compare it to how we human coded a similar brain, and compare the efficiency of both approaches.

The development of this project began with limited token number models. As the token numbers increased, we improved the scope and complexity of the project. 
We developed this project in four "sprints", each of which is described below.

\subsubsection*{Comparison to the dawn of personal computers}

When the author was 12 years old, he got a Commodore Vic 20 for Christmas with 4k ram and a cassette tape memory device. (The family living room TV was the monitor.) He quickly wrote in basic a text-based dungeons and dragons game, where the user was given a choice, e.g. 1) slay the dragon or 2) run! Based on the choice, new scenarios would emerge in the textual story, followed by new choices, and so on. Within about an hour, he coded a few choices and a few dragons and ran out of memory, and the project came abruptly to a screeching halt, never to be looked at again. Who knows how many dragons could have conjured up by his imagination and coded into the game, if only there was more memory than 4k.
This project feels similar. However, we got a lot further this time, but not as far as we want to go. The end goal and limits as of now will be discussed below.

\subsubsection*{Sprint 1: Claude in browser.}
The first sprint used Claude\cite{claude} in a browser (where the code would be copied/pasted from the browser when ready), with some very simple prompts (actual prompts):

\begin{itemize}
    \item \textbf{Prompt:} Write a Python program to send MAVLink commands to a flight controller on a Raspberry Pi. Tell the drone to take off and hover at 50 feet.
    
    \item \textbf{Prompt:} Create a website on the Pi with a button to click to cause the drone to take off and hover.
    
    \item \textbf{Prompt:} Now add some functionality to the webpage. Add a map with the drone location on it. Use the MAVLink GPS messages to place the drone on the map.
    
    \item \textbf{Prompt:} Now add the following functionality to the webpage: the user can click on the map, and the webpage will record the GPS coordinates of the map location where the user clicked. Then it will send a "guided mode" fly-to command over MAVLink to the drone.
    
    \item \textbf{Prompt:} Create a single \texttt{.sh} file to do the entire installation, including creating files and directory structures.
\end{itemize}

The initial prototype was basically functional.
Amazingly, after only the first few prompts, the AI offered this:
\begin{quote}
\textit{
“Would you like me to explain how any part of this works or help you test it? I can also add features like:
}
\begin{itemize}[leftmargin=2em]
    \item \textit{Ability to set altitude for each waypoint}
    \item \textit{Multiple waypoint planning}
    \item \textit{Return-to-home functionality}
    \item \textit{Geofencing boundaries}
\end{itemize}
\end{quote}
This unrequested, volunteered feature demonstrated the model had an intimate knowledge of drones and could suggest additional features based on this knowledge. We were surprised and delighted by this suggestion, which proved immediately the concept and power of AI for robot programming.

The prompt was asked to write a shell script to install all of the files needed onto a fresh Linux instance, including the HTML files, Javascript, Linux Systemd service files, Python code, environment setup, etc.

However, after about a dozen prompts, including several where the model stopped, the conversation came to an abrupt end because of the number of tokens in the model was limited. The actual message was \emph{This conversation has reached its maximum length.}. This was similar to the experience of the author as a 12-year-old and seemed to indicate that a larger project would not be possible with AI.

In order to continue development, we created two new conversations, but both quickly reached the maximum length without fully functioning code with new requested features. The second conversation even created a prompt for the third conversation to start from. Even given this restriction, the model was able to create detailed user manual, complete with troubleshooting guide and safety guidelines!

At the end of sprint 1, the basic prototype website would load but it was not functional enough to be tested in flight. At that point the lack of memory (number of tokens) was an issue that made further development of the project infeasible with that model's limits at that time, using a browser to copy and paste a single install script, so the project was temporarily abandoned. So impressive as it was, further AI development of this project was halted. AI seemed to have reached its limits. 

The unofficial industry standard (more a rule of thumb, not hard and fast) at the time was approximately 1,000 lines of code max for an LLM generated file\cite{hansson2023code,martin2009clean}. 
The install script was 2000 lines of code, so we were clearly at the limit. Furthermore, the install script used multiple languages in one file (python, HTML, Javascript, CSS, bash), which is not an optimum strategy. It was only chosen for simplicity of evaluation in the first phase of the project. The estimated number of man-hours was 16 hours, or two business days. Many (well over half) of those hours were spent waiting for the AI to return a response to the prompt.

\subsubsection*{Sprint 2: Gemini 2.5}
On 25 March 2025, Google announced a public beta of AI programming LLM Gemini 2.5 with 1M tokens. (Claude in sprint 1 was 200k tokens). Excited by the additional capability of the LLM, we started the project again.
We continued with the same in browser, single bash install script approach as in Claude: The prompt was asked to write a shell script to install all of the files needed onto a fresh Linux instance, including the HTML files, Javascript, Linux Systemd service files, Python code, environment setup, etc.

In this sprint, we further guided the LLM to develop functionality and tested it on the bench at each iteration. We used five context windows, each with around 500k tokens (tracked by the UI Gemini Studio). We found after 500k tokens the LLM lost memory of the early parts and necessitated starting a new context window. Each conversation had about 25 individual prompts, some asking for new features, but most reporting a bug in the code during install or on the bench testing which had to be fixed by the LLM.

There were many non-function bona fide bugs identified during this process. The debugging procedure we used was to cut and paste the error message into the LLM and ask it to fix whatever bug caused the error message. We did not have the time to go through the code line by line to try to find the bug manually, nor would we want to, since that goes against the spirit of this project.

One of the largest bugs was getting the syntax of an HTML file inside a bash shell script: The shell script would create an HTML file and write the contents of the HTML file from the script. Surprisingly, this syntax was not correctly coded, and this occurred over and over.

About halfway through this sprint, we had a functional script that would install on a fresh OS install on a Raspberry Pi companion computer on the drone, and we were able to demonstrate it in test flight \#1. However, it still did not properly update the position on the map.

Toward the end of this sprint, the single script install file was still too unwieldy for the LLM, and we asked it to create several individual files (e.g. one HTML file, one CSS file, one python file, etc.) These were cut and pasted into an editor and saved individually, an inefficient process.

Although the history was not recorded precisely, we estimate about 30 total hours of this sprint, mostly sending a new prompt, waiting for the result, cutting and pasting it into Linux, running the shell install script, testing the website to control the drone on the bench, and iterating.

At the end of this sprint, the project was still not fully functional, and was not fully demonstrated in a flight test.

\subsubsection*{Sprint 3: Cursor IDE.}
Recognizing that the project needed multiple files (python , HTML, JavaScript, bash, etc.), we moved to Cursor IDE, a fork of VS Code. This also enabled easy syncing of each version with GitHub. Cursor has its own internal memory locally and can also access various models, including Gemini, ChatGPT, Claude, etc. In addition, having a local IDE allowed testing of basic functionality of the code such as deploying a website on the local machine and local testing of the code Python etc.

About halfway through this sprint, we had a version of the code that was ready for a real flight test. The first fully successful flight test demonstrated the desire functionality: Mode change, takeoff, land, RTL, plot drone on map in real time, click to fly to a position on the map, and arm/disarm.

The process was not without issues. During the process, we tried Claude, ChatGPT, and Gemini. We occasionally switched when one model repeatedly failed on the same task. One simple code change we requested was the ability to specify the IP address of a drone to control as a UI entry on the webpage, rather than hard coded. The challenges here were surprising, and we believe it was because each individual file was still too large and the model did not have the context of all of the files so was unable to track how changes in one segment of code impacted that of other code segments.

Another common challenge was for the LLM to keep track of the dynamics of the drone update information, from the drone, to the back-end, to the front-end, to make the timing efficient, and to handle transient events such as the drone connecting/disconnecting, being in flight and armed vs. on the ground, etc. This indicated that the project had grown beyond the scope of LLM contextual memory.

In order to address this, we asked the LLM to refactor to compartmentalize all the files to smaller-size files. In practice, this generated a lot of extra bugs that did not work. So, it seems from user perspective that it was saturated.

This sprint used about 30 man-hours of prompt engineering and about 35k lines of code written or rewritten "lines of agent", see the appendix, and 51 GitHub commits. 

\subsubsection*{Sprint 4: Windsurf IDE}

In the 4th and final sprint, we switched to Windsurf IDE, a competing fork of VS Code. We hypothesized that it could have more memory capacity awareness of larger codebases vs Cursor. Although both IDEs are proprietary and competitors, Windsurf Cascade was advertised as an agentic experience, ideal for AI-driven multi-file workflows.

We successfully used Windsurf to complete the code refactor, add functionality such as added voice notifications, and add the ability to pick IP address on the website.
This sprint used about 30 man-hours of prompt engineering, about 14k lines of code written or rewritten, and 41 GitHub commits. We also created releases, and had 1.4-1.8 releases on GitHub. A flight test showed enhanced version 1.8 functionality, but the take-off command was not working properly.

A final sprint of 8 hours (and 44 more GitHub commits) in the end led to the full version 2.0. This was comprehensively tested on a desktop/cloud and Raspberry Pi companion computer on the drone, on the bench, and fully tested fully in the air (see below).

\subsubsection*{Adding it up: Man(person) hours, Machine time}

The total \# of Man(person) hours was approximately 100, or about 2.5 weeks. (The detailed statistics available of all the coding are given in the supplementary information.) If AI was instantly responsive, the \# of hours would be about 2 to 4 times less, since a good proportion of the time was spent waiting for the AI to think. As we mentioned above, a vast majority of that time was spent waiting for the AI to reason (while we were checking email, getting coffee, etc.). The total \# of GitHub commits was 120. The total number of lines of code for the project (version 2.0) is around 10k.

\subsubsection*{Comparison to Cloudstation}
This project of code is almost identical to a similar project we developed over the last four years called called \emph{Cloudstation}\cite{cloudstation,cloudstation2024}
That project had four major releases. The first release was the result of six undergraduates working for two quarters. The second release was the result of three graduate students working for two quarters. The third release was the results of one graduate student working for 2 quarters. The fourth release was the result of the author working for maybe 1-2 weeks. If we estimate the class credit as 3 hours of contact time and 6 hours of work per student per week, the math comes to about 2000 hours. (One quarter = 10 weeks, so 90 hours). Thus, this paper demonstrates approximately 20x less hours of work for a very similar result.

\subsection*{Architecture}

\begin{figure*}
    \centering
    \includegraphics[width=127mm]{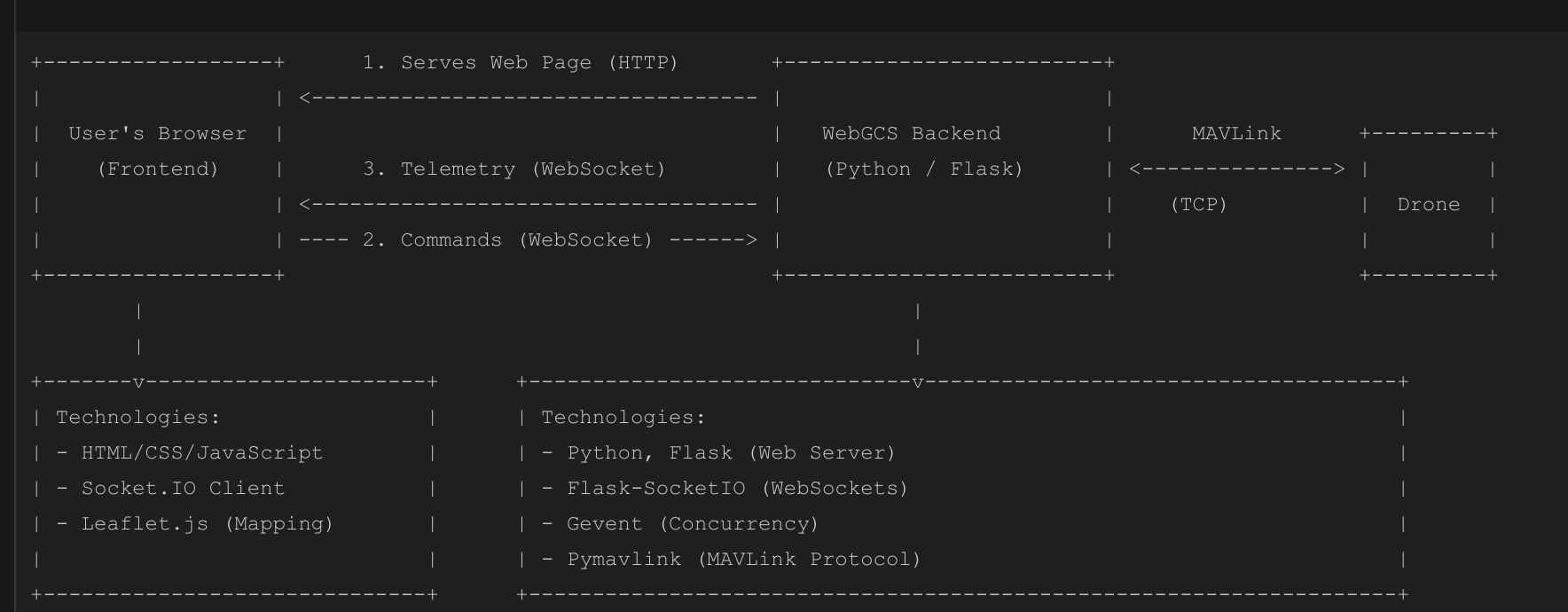}
    \caption{\textbf{Architecture.}
    The modules and how they interact. 
    }
    \label{fig: architecture}
\end{figure*}

In Figure~\ref{fig: architecture} we show the architecture. The webserver is based on Flask, a Python based architecture. The most well developed and documented library for MAVLink is PyMavlink, so the integration is easiest with this engine. WebSocket is used for real time communication between the modules.

\subsubsection*{AI designed the architecture}
The choice of architecture was entirely from the AI. No human input was given on how to structure the project.

\subsubsection*{AI described the architecture}
The figure was generated by AI with a simple prompt \emph{"can you make a diagram of the architecture"}. The AI was asked to create an image that describes the architecture. Thus, AI has internal knowledge of the architecture and how all software modules interact, and also how the drone in the air and the pilot on the ground interact with the architecture.

\subsubsection*{Assessment of AI decision on architecture}

The choice of architecture is excellent. The only disadvantage is that Flask is not meant to scale to production of thousands of drones and thousands of instances of WebGCS. However, it was not asked to do that. It was asked to design something for a single drone. If we had to do it ourselves, we would have chosen the same thing.

We speak from experience. We have recently published a similar project, written by a team of six undergraduates over the course of a year, followed by improvement of 3 masters students over the course of the following year, followed by another several months.  That architecture is described in detail in~\cite{cloudstation,cloudstation2024}. Here we will briefly compare it to the AI designed architecture. Cloudstation also used Python, but did so with Django. This was more complex to code but otherwise equivalent to Flask for this project for single drones. Django had the advantage of a log-in credentialing system, which this AI was not asked (or intended) to do. Cloudstation also used a database to store drone state information, with the idea of scaling to drone swarms in the future. Cloudstation used Leaflet instead of OpenStreetMap for the map API. Both are equivalent for the purposes of this project.
Finally, Cloudstation ran in a Docker container, which was just an alternative and complementary way of managing all packages.

There are several other proprietary architectures that achieve the same cloud-based web GCS systems, which we reviewed in detail in ref.~\cite{cloudstation}. Since they are proprietary, we have no way to know the internal architecture for comparison.

In summary, based on years of experience designing and implementing similar projects, our assessment is that the AI designed architecture is the most appropriate choice for this project (single drone). 

\subsection*{Pilot interface}

\begin{figure*}
    \centering
    \includegraphics[width=185mm]{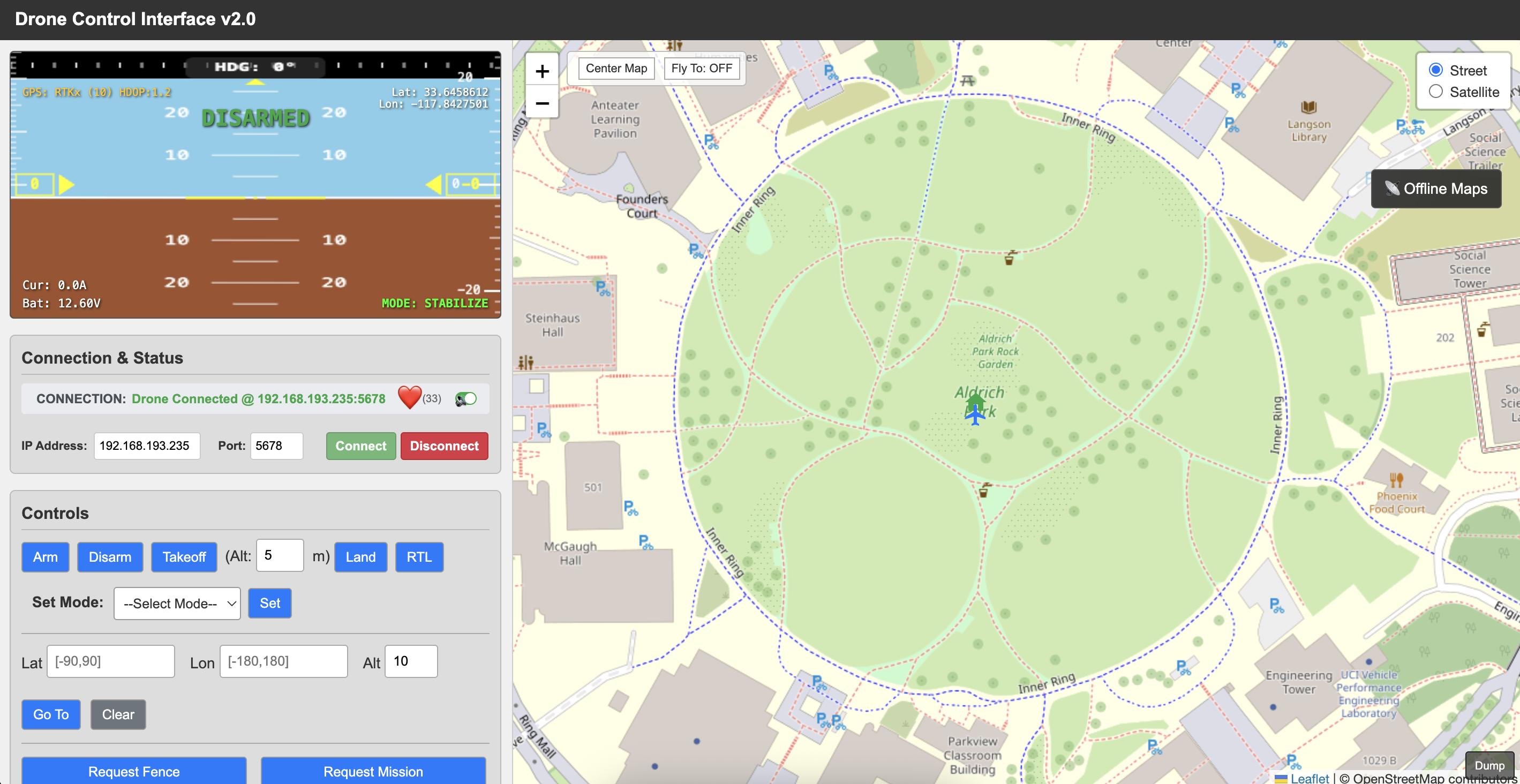}
    \caption{\textbf{Website.}
    The drone location on the map is updated in real time. The HUD on left gives drone information, similar to other GCS software. Control buttons initiate various autonomous actions such as takeoff/land/return to home.
    }
    \label{fig: WebGCSUI}
\end{figure*}

The pilot interface was fully coded by AI. The AI was told to have a map and buttons, and little else. The AI autonomously chose to implement a heads-up display without any specific prompt to do so. We asked the AI to make some cosmetic modifications to the UI and add some debugging features such as logging messages, which it did. The current version graphical user interface (GUI) is shown in Figure~\ref{fig: WebGCSUI}.

\subsection*{Performance: Flight tests}

\begin{figure*}
    \centering
    \includegraphics[width=185mm]{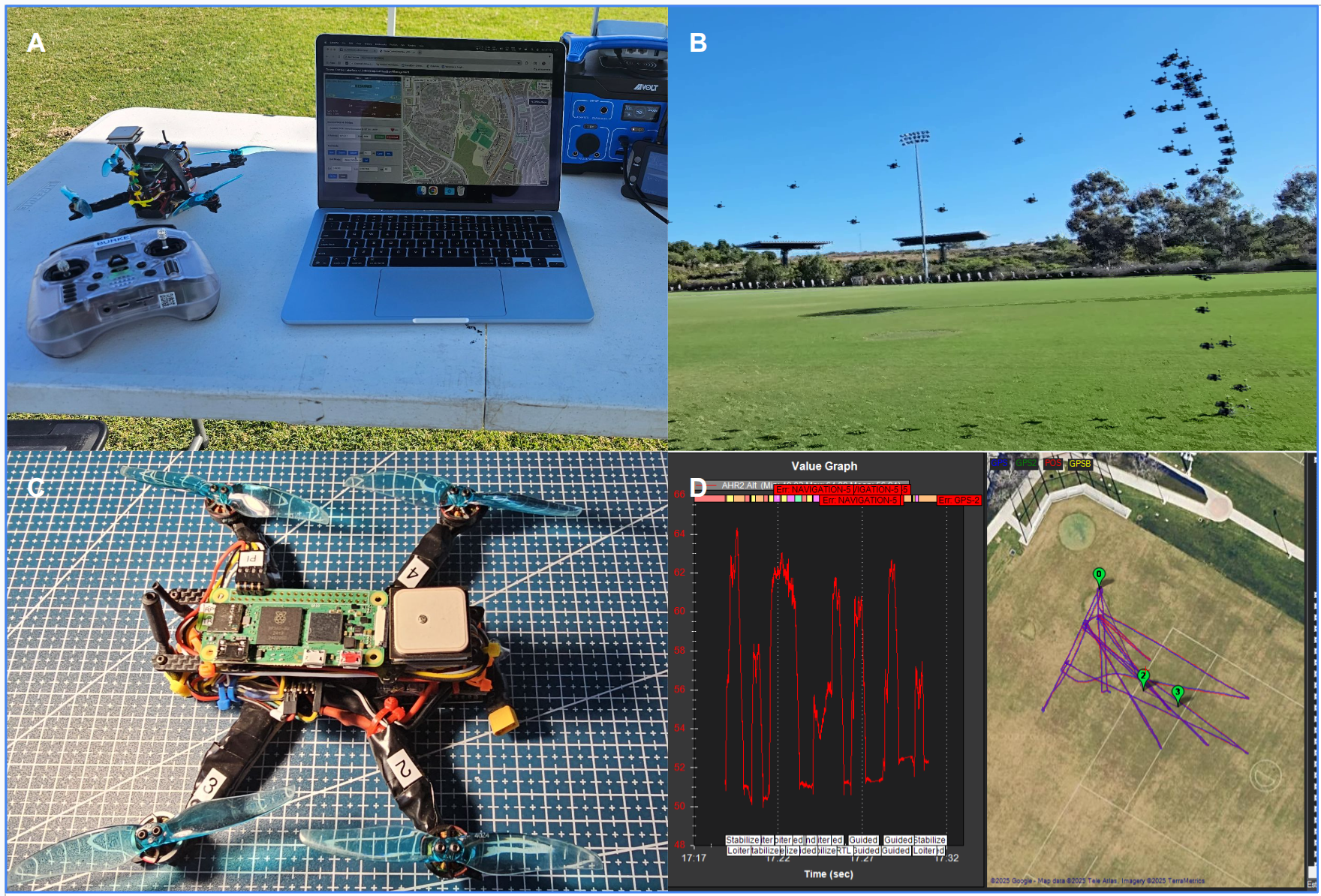}
    \caption{\textbf{Flight demonstrations.}
    (\textbf{A}) Website seen on the laptop was hosted on the drone. The small handheld radio is a backup and can be used to manually pilot the drone if the WebGCS fails for any reason at any time during the flight.
    (\textbf{B}) Timelapse image of a flight where only the WebGCS was used for drone control.
    (\textbf{C}) Picture of the drone. The Raspberry Pi Zero 2 W on the top of the drone hosted the AI generated code.
    (\textbf{D}) Post flight log analysis showing the drone flight path. The altitude vs. time shows multiple flights, and the map shows the location of the drone during the flights.
    }
    \label{fig: Flightdemo}
\end{figure*}

Flight tests were performed with a small drone with a Raspberry Pi (Figure~\ref{fig: Flightdemo}). Details of the drone are provided in methods and ref.\cite{daa}. The WebGCS was hosted on the drone, and a laptop was logged onto the drone WiFi hotspot. The connection was stable up to 100 m away, the maximum distance tested.

The flight tests were flawless, with a few bugs. One of the times, the drone did not update its position on the map. Another time it did not take off when commanded. As of version 2.0 there are no known bugs.

The successful flight test consisted of the following sequence of events (Figure~\ref{fig: Flightdemo}), using the website to trigger each event:
\begin{itemize}
    \item Arm
    \item Takeoff
    \item Fly to a point on the map
    \item Return to launch
\end{itemize}

Several test flights are shown in Figure~\ref{fig: Flightdemo}D, with gradually increasing level of control from WebGCS interface only. During the final test flights, the backup remote control was not used at all. The entire flight was controlled by the AI coded brain of the drone in the air and the website interface.

Although these missions were quite modest and simple, they demonstrated proof of concept and operation. However, this method is easily scalable to much more sophisticated and longer distance missions.

\subsubsection*{Safety protocols}
Using AI-generated code without human checking for bugs to control a drone in the air is a dangerous thing to do. When human safety is involved with AI generated code that has not been audited by a human, great care must be taken. In this work, we flew the drone only over a large field without humans around. We also had a standby remote control with redundant manual control ability if needed. Finally, we implemented a strict geofence in the software.

\subsection*{Performance: Simulation and deployment in the cloud}

\begin{figure*}
    \centering
    \includegraphics[width=88mm]{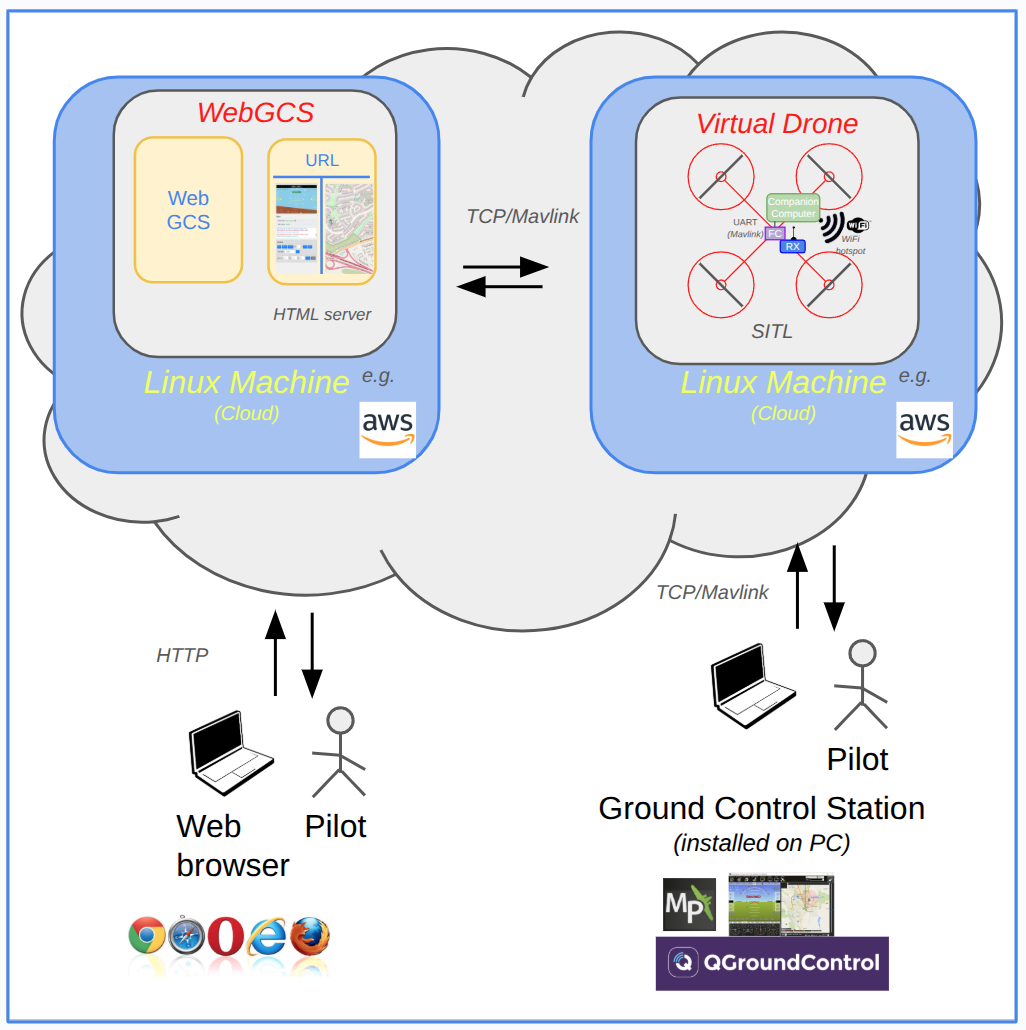}
    \caption{\textbf{Virtual drone.}
    One (cloud based) Linux machine hosts a virtual drone, which can be control with Mavlink/TCP from anywhere in the world with an internet connection. The bottom right pilot logs on using a PC with a ground control station installed on it, such as Mission Planner, or QGroundControl. Another Linux machine (also cloud based) hosts WebGCS, which connects to the virtual drone also through TCP/Mavlink. The pilot on the bottom left only uses a web browser to log into WebGCS and control the virtual drone.  
    }
    \label{fig: VirtualDrone}
\end{figure*}

Ardupilot has a mature drone simulator program that includes physics\cite{ardupilotSITL}. We have developed an install script to set it up in a Linux cloud instance and run as an "always on" service : A virtual drone in the cloud\cite{createsitlenv}. Without firewalls, any pilot can fly the virtual drone, using the IP address of the cloud Linux instance to connect. The pilot would need Mission Planner or QGroundControl installed on their local PC. With a firewall, a good option is to create a VPN such as Zero Tier. In addition, WebGCS can be installed on \emph{another} Linux instance in the cloud. Again, with this setup, the drone pilot can log onto the WebGCS server, this time using only a web browser, without the need for a local installation of any other software. In the WebGCS interface, the IP address of the virtual drone can be entered, allowing testing of using WebGCS to fly the virtual drone, all without the need for any software on the local machine. This is an excellent way to test the program in a safe way. Detailed video and text tutorials that walk the user step by step through the process are available on YouTube\cite{virtualdrone_video,lockdown_virtualdrone} and GitHub\cite{createsitlenv}. Figure~\ref{fig: VirtualDrone} shows a typical simulated drone test architecture.

\section*{DISCUSSION}

\subsection*{Scaling and limits}

What is the relationship between the number of tokens in an LLM AI model, and the number of lines of code that model can write, maintain, refactor, enhance, and debug? To our knowledge, there is (as yet) no general answer to this question\cite{chen2021evaluating,fan2023large,beer2024examination,bogomolov2024long,hua2025researchcodebench,assogba2024evaluating, hsieh2024ruler, zhang2024bench, yen2025helmet}. We discuss the practical experience from this paper, and its implications for the future of AI in robotics below.

\subsubsection*{This work in practice: 10k lines of code}

In this work, our codebase ended up at about 10k lines of code. This is already quite an accomplishment for an AI code project written completely from scratch with no human lines of code. In our experience, even the best IDEs, LLMs, and models were unable to completely have the entire codebase in memory context sufficient for any given task. For small code changes this was not an issue, for example to change the color of a button. But code changes which required knowledge of the plumbing of the flow of information were more troublesome. In the drone case, the flow and storage of information is one of the primary tasks of the code, and it flews through many modules from the drone, to the computer, to the back end, to the front end, and back again when a pilot requests an action, such as takeoff or land. In our experience, this is about as complex of a codebase as modern AI models can handle, as of the writing of this paper.

\subsubsection*{Comparison of our work to state of the art in lines of code}
Although it depends on the codebase properties, for the purposes of this discussion, we assume 10 tokens as equivalent one line of code. For large context models, for example,  Rando\cite{rando2025longcodebench} found \emph{The accuracy of Claude 3.5 Sonnet on LongSWE-Bench drops from 29\% to 3\% when increasing context length from 32K to 256K.} Claude Sonnet is a model marketed for its long-context prowess.
 As our codebase is around 10k lines of code, we are right at the edge of the scaling of current models. This paper by Rando was not read until after our experience with this project, and so our experience is an unbiased, independent (single case) real world example that is consistent with the more general findings of Rando et al\cite{rando2025longcodebench}.

\subsubsection*{Feasible next steps: Going up and down the tech stack}

A feasible next step would be to extend AI coding up the tech stack to higher order reasoning, such as autonomous navigation based on broad goals. ROS2 in some sense is one of the human written version of that: not just GCS but true autonomy. It is similar to but different from the AI beating the human in a drone race\cite{kaufmann2023champion}, but coding it all with AI \emph{itself}.

Another feasible next step would be using AI coding to go down the tech stack. For example, could one list the requirements of the low level flight control software such as Ardupilot or PX4 as a technical specification and vibe code a new version of Ardupilot? Since Ardupilot codebase has about 1M lines of code, most likely the next step should be some specific task, such as racing, demonstrated by\cite{kaufmann2023champion}.

\subsubsection*{Feasible next steps: AI agent swarms controlling real drone swarms}

The next logical steps could be AI agent swarms controlling real drone swarms, or light shows\cite{jin2024llms}. This is possible but beyond the scope of this work.

\subsubsection*{This work taken to the ideal extreme: 1M+ lines of code}

The GCS created here had basic functionality: Plot the drone on a map, provide detailed telemetry data such as airspeed, altitude, orientation on a heads up display, and provide button-based control of autonomous actions such as land, takeoff, etc. However, the more complete GCS systems have much richer feature sets, although they are not website based. One goal to extend this work would be to list all the features of Mission Planner or QGroundControl, feed that into an AI prompt, and ask the AI to make a web based version of those programs. Based on this work, this would not be possible with the approach provided herein. Perhaps if the recent move towards AI agents swarms is applied to this problem, it could be tackled, although how to do that is not clear. The Mission Planner codebase has 1.7M lines of code. QGroundControl has 0.9M lines of code. The Linux kernel as of this writing has 40M lines of code. However, AI models are advancing extremely rapidly. When will AI be able to write something like the entire Linux OS from scratch, and after that feat, what will they next be capable of?

\subsection*{Lessons learned}

The following practical lessons were learned:

\begin{enumerate}
    \item Use an IDE. This is essential, especially as the codebase grows and the project involves more than one language.
    \item Use GitHub. Version control is critical in AI development, especially as AI can and often does break functionality. Being able to revert to a previous version reduces the risk of catastrophic code failure created by the AI that the human cannot understand. Make commits often. The AI IDEs can actually make the commits on command and generate comments describing the new features since the last commit.
    \item Be specific in asks. The more specific the feature request, the more likely the AI is to implement it as desired. This is the direction the industry is moving, from \emph{vibe coding} to \emph{spec coding}, where the detailed specification is listed out prior to requesting the code be written.
\end{enumerate}

\subsection*{Request for industry}

When trying to track the detailed commits, code evolution, and prompt history, we were surprised to find that industry does not have a standard, straightforward method to save a log of the history under a given account. In fact, some companies delete the history after a certain period of time, and there seems to be no public facing industry standard about data retention for consumers. Of the total history, several tracking metrics would be useful to have integrated into the models: Track hours, track number of prompts, exact prompts, number of tokens. A challenge is that the models and even number of tokens are considered proprietary, closely guarded trade secrets, so research in this area is difficult to progress (e.g. number of tokens vs. number of lines of code).

\subsection*{Safety}
 
The general problem of testing AI generated code is an open problem and was not solved comprehensively or rigorously in this work. While this work demonstrates the importance of such an advance in a visceral way, it is beyond the scope of this paper.

Nevertheless, we were able to test the features in a relatively safe environment. Testing it in a simulation environment is a good approach, but when dealing with the physical world, it may deviate from simulations. At this point it is not clear what the correct testing procedure should be if, for example, this is deployed on drones in higher risk operations, for example operations over people, or beyond visual line of sight. While human code can be debugged and tested line by line, more research needs to be done on the safety and reliability of AI generated code.

\subsection*{Conclusion}

In conclusion, we have demonstrated a robot making a robot’s brain. While we hope the outcome of Terminator never occurs, this work demonstrates the capability of AI to create new versions of itself in the realm of drones. In our work, a redundant RC transmitter under human control is always available as a backup in case manual override is needed. That is how we implement human control of autonomous robots in this case. How to keep Terminator from happening to humanity is a future research topic, beyond the scope of this work.

\section*{MATERIALS AND METHODS}
\label{sec: Methods}

\subsection*{Software}
The following models were used: Google Gemini 2.5. ChatGPT 4.0. Claude Sonnet 3.5, 3.7. The following IDEs were used: VS Code, Cursor, Windsurf. The code was developed primarily on an AMD Linux laptop, an Intel i9 desktop, and a Macbook Air M4. The drone firmware was Ardupilot, ArduCopter version 2.6.

\subsection*{Hardware}
The drone was a sub-250g 4" drone with ELRS radio connections. The drone was powered by a 2S LiPo battery.
A Raspberry Pi Zero 2 W was connected to one of the UARTs of the flight controller (Matek F405 Wing, powered by an STM32 F4 based microcontroller) and also powered by one of the 5V BECs on the flight controller. The total current of the avionics at idle throttle was 0.45 A. About 0.1 A of this was for the Raspberry Pi. Details of the build (bill of materials) are in~\cite{ucidrone} and~\cite{daa}.

\subsection*{AI}

Obviously AI was used to write the code. The only other place AI was used in this work was in the generation of the robot image in Figure~\ref{fig: manvsmachine}B, the generation of the code architecture in Figure~\ref{fig: architecture}, and to fine tune the abstract verbiage from a draft abstract. Other than that all work including writing this manuscript was done by a human, line by line, word by word, the old fashioned way.

\bibliographystyle{sciencemag}
\bibliography{references.bib}

\section*{}
\textbf{Acknowledgments:} We thank Shanyu Mou and Bogdan Kovtun for flight spotting.
\textbf{Competing interests:} None identified. \textbf{Data and Materials Availability:} All (other) data needed to evaluate the conclusions in this paper are present in the paper or Supplementary Materials. The data and code for this study have been deposited in GitHub (see references).

\clearpage

\section*{Supplementary Materials}

\begin{center}
\includegraphics[width=7in]{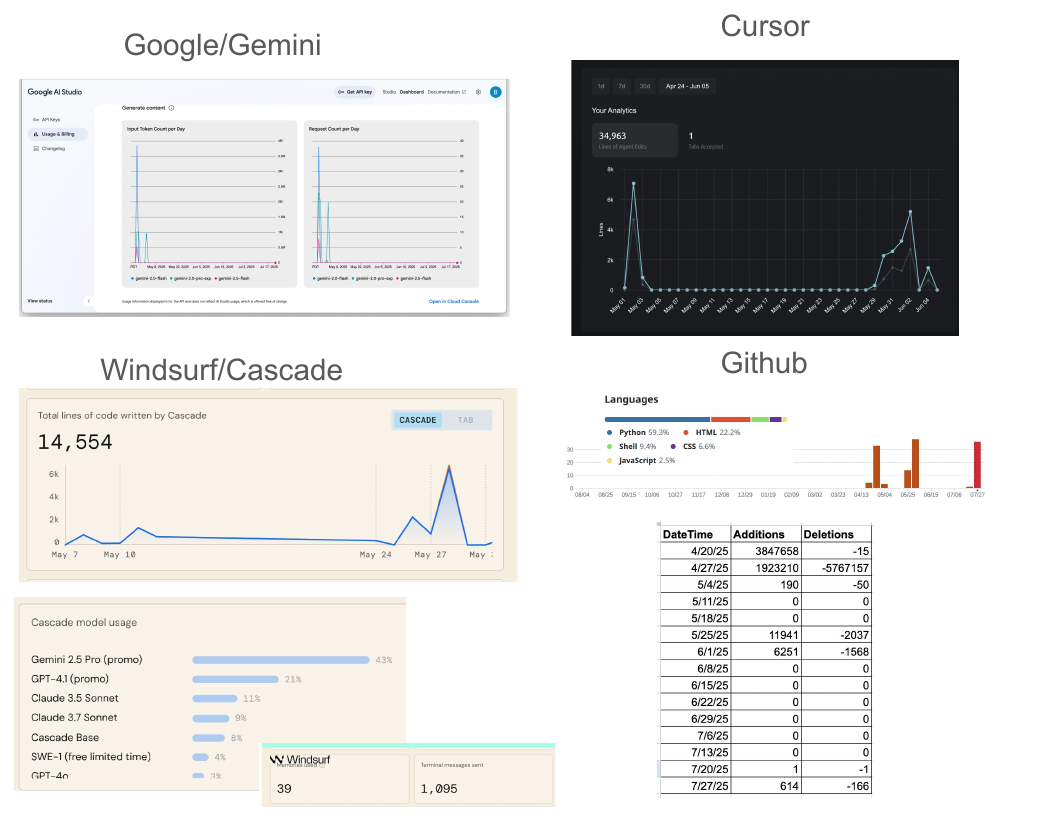}
\end{center}

\noindent\textbf{Figure S1.} \textbf{Statistics.} A) The usage statistics for the models, when available, and in the format provided by the company.

\end{document}